\documentclass[sn-mathphys]{sn-jnl}

\usepackage{natbib}
\usepackage{url}
\usepackage{xcolor}
\usepackage{hyperref}
\usepackage{ulem}

\jyear{2021}%

\theoremstyle{thmstyleone}%
\theoremstyle{thmstyletwo}%

\theoremstyle{thmstylethree}%

\begin{document}
	
	\title[LiteLSTM Architecture Based on Weights Sharing for Recurrent Neural Networks]{LiteLSTM Architecture Based on Weights Sharing for Recurrent Neural Networks}
	
	
	\author*[1]{\fnm{Nelly} \sur{Elsayed}}\email{elsayeny@ucmail.uc.edu}
	
	\author[1]{\fnm{Zag} \sur{ElSayed}}\email{elsayezs@ucmail.uc.edu}
	
	\author[2]{\fnm{Anthony S.} \sur{Maida}}\email{maida@louisiana.edu}
	
	\affil*[1]{\orgdiv{School of Information Tecchnology}, \orgname{University of Cincinnati}, \orgaddress{\street{2610 University Cir}, \city{Cincinnati}, \postcode{45221}, \state{Ohio}, \country{United States}}}
	
	\affil[2]{\orgdiv{School of Computing and Informatics}, \orgname{University of Louisiana at Lafayette}, \orgaddress{\street{301 E. Lewis Street}, \city{Lafayette}, \postcode{70503}, \state{Louisiana}, \country{United States}}}
	
	
	
	\abstract{Long short-term memory (LSTM) is one of the robust recurrent neural network architectures for learning sequential data. However, it requires considerable computational power to learn and implement both software and hardware aspects. This paper proposed a novel LiteLSTM architecture based on reducing the LSTM computation components via the weights sharing concept to reduce the overall architecture computation cost and maintain the architecture performance. The proposed LiteLSTM can be significant for processing large data where time-consuming is crucial while hardware resources are limited, such as the security of IoT devices and medical data processing. The proposed model was evaluated and tested empirically on three different datasets from the computer vision, cybersecurity, speech emotion recognition domains. The proposed LiteLSTM has comparable accuracy to the other state-of-the-art recurrent architecture while using a smaller computation budget.}

	\keywords{LiteLSTM, weights sharing, LSTM, recurrent neural networks, IoT, MNIST}
	
	
	
	\maketitle
	
	\section{Introduction}\label{sec:introduction}

	%
	%
	%
	%
	Sequential data modeling such as text, univariate and multivariate time series, audio signals, biological signals, spatiotemporal sequences (videos), amino acid amd genetic sequences requires an apparatus that can recognize the temporal dependencies and relationships within the sequential data.
	In the early 1980s, the recurrent neural network (RNN) was designed as the first neural network approach that targeted sequential data problems~\cite{bourlard1989speech,siegelmann1995recurrent,deepLearnigBook}. 
	The RNN architecture can capture temporal dependencies due to the sense that it recursively integrates the current new input into its self-previous output~\cite{graves2009novel}. Since it has an unrestricted but fading memory for the past, it can employ the temporal dependencies to influence the learning of the structure within the data sequences~\cite{elsayed2019gateddissertation}.
	The RNN has been applied in different research areas such as handwriting recognition~\cite{graves2009novel,stuner2020handwriting,carbune2020fast}, speech recognition~\cite{sak2014long,Graves2013SpeechRW,zeyer2017comprehensive}, language modeling~\cite{mikolov2010recurrent,mikolov2011extensions,sundermeyer2012lstm}, machine translation~\cite{ren2020use,bridle1990alpha,bahdanau2014neural}, action recognition~\cite{du2015hierarchical,ullah2017action,adewopo2022baby}, accident recognition~\cite{bortnikov2019accident,adewopo2022review,fatima2021global}, stock prediction~\cite{kamijo1990stock,elsayed2021intrusion,azumah2021deep}, video classification~\cite{yang2017tensor,ogawa2018favorite}, intrusion detection systems~\cite{debar1992application}, time series prediction~\cite{han2004prediction}, and mental disorder prediction~\cite{petrosian2001recurrent}. 
	
	However, the RNN has a significant weakness: its ability to learn long-term dependencies is limited due to the vanishing/exploding gradient problem. There are several attempts to solve the RNN major design problem and enhance its overall performance, as the RNN loses the ability to learn when the error gradient is corrupted. To solve the vanishing/exploding gradient, extensions to the RNN architecture require adding an internal state (memory) that enforces a constant error flow through the RNN architecture stage. This constant error flow enhances the robustness of the error gradient over longer time scales. In addition, a gated control over the content of this internal state (memory) is also needed~\cite{hochreiter1997a}.
	
	Nevertheless, this early LSTM model had significant weaknesses. When it was early designed by Hochreiter and Schmidhuber~\cite{hochreiter1997a}, the LSTM model input data was assumed to be prior segmented into subsequences with explicitly marked ends that the memory could reset between each irreverent subsequences processing~\cite{hochreiter1997a,gers2000learning}. Moreover, this LSTM architecture did not have an internal reset component in case of processing continual input streams. Therefore, when the LSTM processes continuous input streams, the state action may grow infinitely and ultimately cause the LSTM architecture to fail~\cite{gers2000learning}.
	
	In 2000,~\cite{gers2000learning} proposed a solution for the original LSTM problem that was proposed in~\cite{hochreiter1997a}.~\cite{gers2000learning} added a forget gate beside the input and output gates into the LSTM architecture that resets the LSTM memory when the input is diversely different from the memory content and helps to remove the unnecessary information that the LSTM memory carries through the time. This LSTM approach~\cite{gers2000learning} is widely used to solve various problems such as speech recognition~\cite{sak2014long,soltau2016neural,chorowski2014end,miao2015eesen,graves2013hybrid}, language modeling~\cite{sundermeyer2012lstm,merity2017regularizing,sutskever2014sequence,miyamoto2016gated}, machine translation~\cite{cho2014properties,bahdanau2014neural,luong2014addressing,luong2015stanford}, time series classification~\cite{karim2018lstm,karim2018multivariate}, image segmentation~\cite{stollenga2015parallel,chen2018deeplab,reiter2006combined}, and video prediction~\cite{cho2014properties}.
	
	However, this model also has pivotal weaknesses. First, the architecture does not have a direct connection from the memory state to the forget, input, and output gates. Hence, there is no control from the memory to the gates that could assist in preventing the gradient from vanishing or exploding. 
	Second, the Constant Error Carousel (CEC) does not have influential conduct over the forget and input gates when the output gate is closed (i.e. the output gate produces zero value output), which could negatively affect the model due to the lack of primary information flow within the model~\cite{gers2002learning,gers2000recurrent}.
	
	To handle these problems in the standard LSTM, in 2002,~\cite{gers2002learning} added the peephole connections from the memory state cell to each of the LSTM forget, input, and output gates.
	The peephole connections allowed the memory state to exert some control over the gates, reinforcing the LSTM architecture and preventing the lack of information flow through the model during the situation that leads to the output gate being closed~\cite{gers2002learning}.
	
	The peephole added a generalization element to the standard LSTM~\cite{elsayed2020reduced}. However, the major weakness of this architecture is that it becomes cost expensive due to the significant increase in the number of trainable parameters, memory, processing, and storage requirements to train the model and save the trained weights of the model and training time.
	
	However, there is still growing interest in studying and applying the LSTM architecture to solve various sequential problems in different research domains due to the LSTM outperforming the GRU in several tasks when problems have large training datasets~\cite{greff2017lstm}. Moreover, Greff et al.~\cite{greff2017lstm} proposed research in 2017 showed that the LSTM exceeds the GRU performance in language modeling-related tasks. On the other hand, in some problems where the training datasets are small, the GRU outperforms the LSTM using a smaller computation budget~\cite{chung2014empirical}.
	
	As the era of big data requires robust tools to manipulate large data processing. In addition, it requires accelerated, time-consuming tools to process the data. Moreover, as the world tries to reduce the Carbon (CO2) footprint~\cite{bocken2012strategies} by reducing the usage of high-performance hardware~\cite{calza2017types,zaghloul2021green,al2012green,elsayed2021autonomous}, the LSTM implementation requirements cost is considered one of the significant LSTM drawbacks.
	
	Spatiotemporal prediction problems are challenging to solve, utilizing only a gated recurrent architecture. Implementing such models is quite expensive from both resources and value aspects as a large number of parameters, rapid processors, large processing memory, and memory storage are needed. In addition, such models demand considerable time to train, validate and test. Moreover, implementing such a model for real-time training is a challenge.

	This paper attempts to evolve several computational aspects into a sophisticated performance level. This paper proposed a novel recurrent gated architecture using one gate: Lite Long Short-Term Memory (LiteLSTM). The proposed LiteLSTM employed the concept of sharing weight among the gates introduced in the GRU~\cite{chung2014empirical} to reduce the model computation budget. Also, it employs memory control over the gate using the peephole connection over the one gate. Beside
	Compared to the LSTM, Peephole LSTM, and GRU, the LiteLSTM has a smaller computation budget and implementation requirements, maintaining comparable accuracy. Due to its smaller computation budget, the LiteLSTM has a significant training time reduction compared to the LSTM. That allows the LiteLSTM to be implemented without a CO2 footprint requirement. 
	
	This paper is organized as follows: Section~\ref{DeepRNNS} provides a brief overview of the RNN, standard LSTM, peephole LSTM, and GRU architectures. Section~\ref{LiteLSTM} provides the LiteLSTM architecture design concept details, Section~\ref{emperical} shows empirical results for LiteLSTM implementation on three applications from three different research domains: computer vision (using MNIST~\cite{lecun1998mnist}, cybersecurity anomaly detection in IoT (IEEE IoT Network Intrusion Dataset)~\cite{q70p-q449-19}, and speech emotion recognition (TESS dataset~\cite{dupuis2010toronto}). 
	\begin{figure*}
		\centering
		\includegraphics[width=12cm,height=3.5cm]{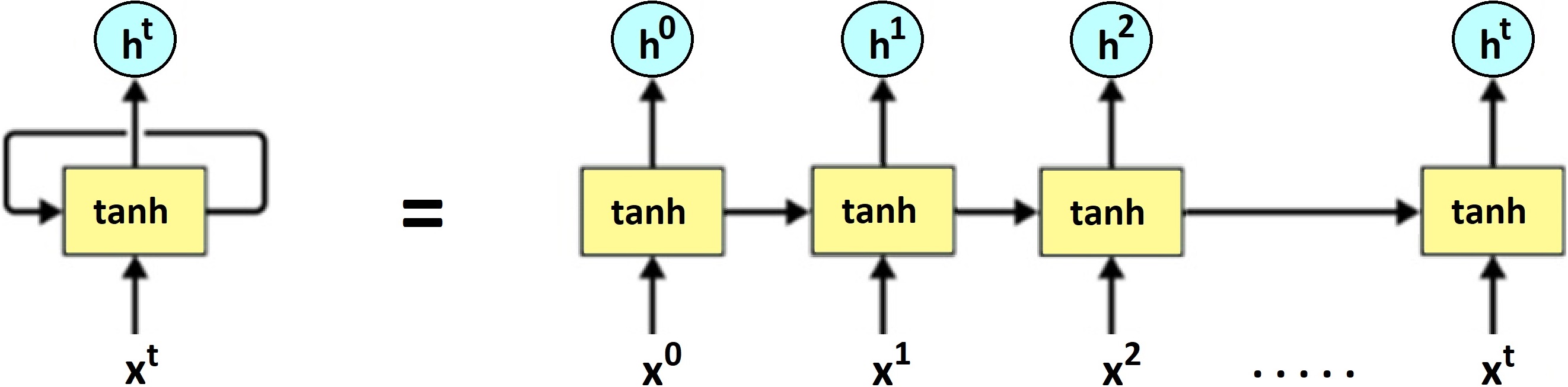}
		\caption{The RNN basic architecture and its corresponding unfolded in time representation~\cite{rnn_diagram_blog}.}
		\label{rnn_architecture}
	\end{figure*}
	
	\section{Recurrent Neural Networks}\label{DeepRNNS}
	
	\subsection{Basic RNN Architecture}
	The recurrent neural network (RNN) basic architecture is shown in Figure~\ref{rnn_architecture}. The left diagram shows the RNN architecture. The unfolded (unrolled) in time RNN representation is shown in the right diagram starting from the time step $0$ to time step $t$. The RNN is transformed into a feedforward network that can be trained by backpropagation. This algorithm is called backpropagation through time (BPTT)~\cite{werbos1990backpropagation}. The RNN feeds its previous output vector $h^{(t-1)}$ at time step $t-1$vand the current input vector $x^{(t)}$ to calculate the RNN output $h^{(t)}$ at the current time step $t$. This method allows the RNN to identify and utilize temporal information to influence learning in the data sequences. 
	
	The basic RNN suffers from the vanishing/exploding gradient problem~\cite{ceni1807interpreting}, limiting the model's ability to learn long-term dependencies within the sequential data. This is because the RNN does not have any element in its architecture design components that could maintain a constant error flow through the recurrent model. The principle of adding gates as supporting components into the recurrent architecture was proposed to solve this problem.

	At a given discrete time step $t$, the RNN output is calculated as follows:
	\begin{equation}\label{eqn:rnn}
		h^{(t)}= \mathrm{tanh}(W x^{(t)} + U h^{(t-1)}+b)
	\end{equation}
	where $x^{(t)}$ is the RNN input at time step $t$. The $h^{(t)}$ and $h^{(t-1)}$ are the RNN outputs at time steps $t$ and $t-1$. The feedforward and recurrent weights are represented by $W$ and $U$, respectively. The weights are shared across time steps. $b$ is the RNN model bias. 
	
	\subsection{Standard Long Short-Term Memory (LSTM)}

	\begin{figure}
		\centering 
		\includegraphics[width=8cm,height=5cm]{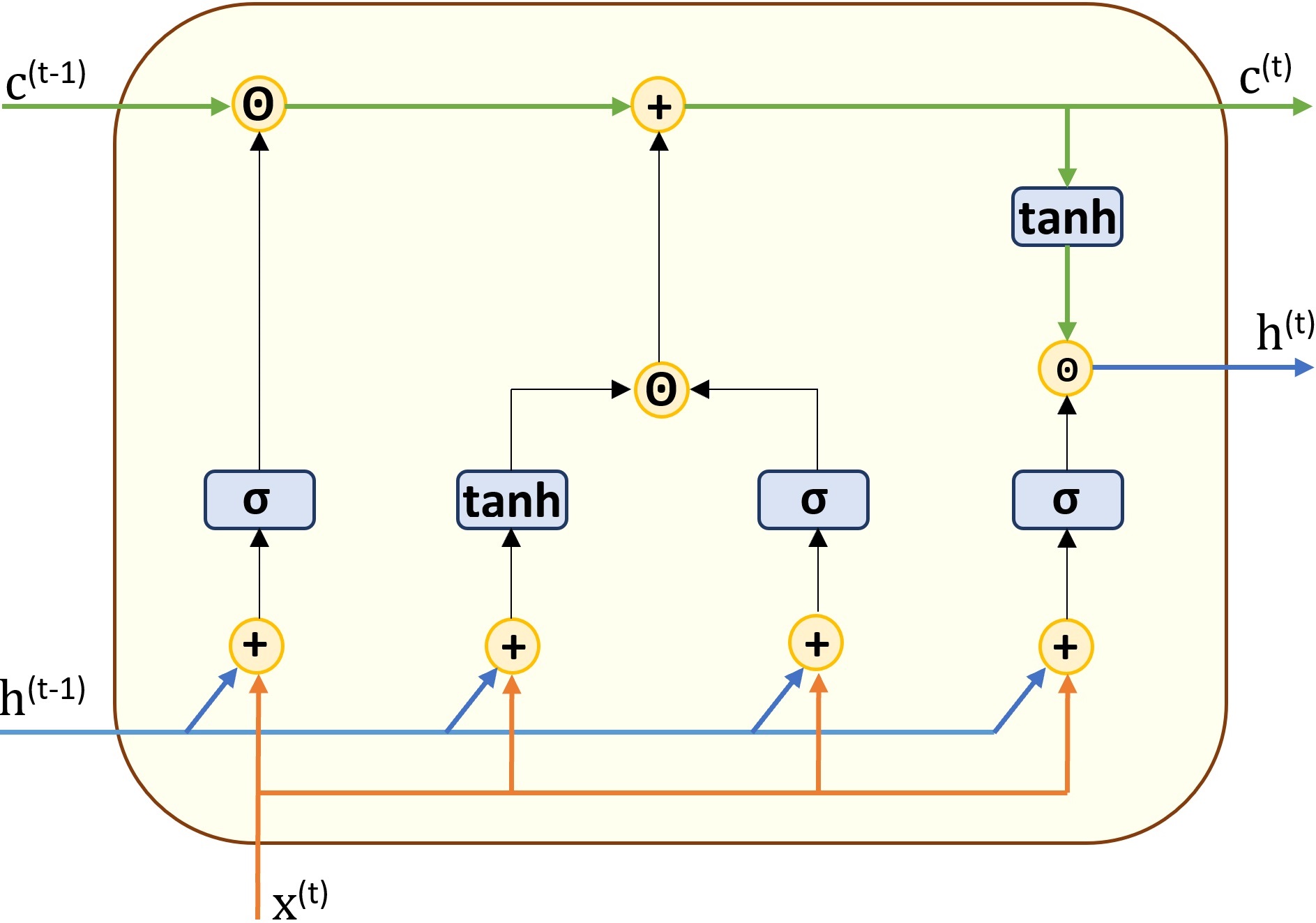}
		\caption{The standard LSTM unrolled architecture.}
		\label{figUnitArchitecture}
	\end{figure}
	
	Gers et al.~\cite{gers2000learning} proposed the standard LSTM architecture in 2000 as an improved version of the first LSTM architecture, which was proposed in 1997 by Hochreiter et al.~\cite{hochreiter1997a}. This standard LSTM aimed to solve the continuous input stream problem, which allowed the memory state cell values to grow in an unbounded fashion, causing saturation of the output squashing (activation) function. Gers et al.~\cite{gers2000learning} proposed to add an additional gate to the LSTM architecture: forget gate $f$ to reset the LSTM memory when the input is diversely different from the memory content and serves to remove the unnecessarily information that the LSTM memory holds through time.
	
	Figure~\ref{figUnitArchitecture} shows the standard LSTM unfolded architecture where $c^{(t)}$, $h^{(t)}$ are the memory state cell and LSTM output at time $t$, respectively. The symbol $\odot$ denotes the element-wise (Hadamard) multiplication~\cite{gers2000learning,elsayed2019reduced} and $\sigma$ denotes the logistic sigmoid function. $b_i$, $b_g$, $b_f$, and $b_o$ are the biases of each gate. $W$'s are the feedforward weights and $U$'s are the recurrent weights. 
	
	The value of each component in the standard LSTM is calculated as follows:
	\begin{align}
		i^{(t)}&= \sigma(W_{xi} x^{(t)} +U_{hi}  h^{(t-1)}+ b_i)\label{eqn:i_lstm}\\ 
		g^{(t)}&= \mathrm{tanh}(W_{xg} x^{(t)} +U_{hg}  h^{(t-1)}+ b_g)\label{eqn:g_lstm}\\ 
		f^{(t)}&=\sigma(W_{xf} x^{(t)} +U_{hf}  h^{(t-1)} + b_f)\label{eqn:f_lstm}\\ 
		o^{(t)}&=\sigma(W_{xo} x^{(t)} +U_{ho}  h^{(t-1)} + b_o)\label{eqn:o_lstm}\\ 
		c^{(t)}&= f^{(t)}\odot c^{(t-1)} + i^{(t)} \odot g^{(t)}\label{eqn:s_lstm}\\
		h^{(t)}&= \mathrm{tanh}(c^{(t)})\odot q^{(t)}\label{eqn:h_lstm}
	\end{align}
	where $i^{(t)}$, $f^{(t)}$, and $o^{(t)}$ are the input, forget, and output gates, respectively. 
	The gates are constrained to have activation values between zero and one to indicate their status: open, closed, partially open, or partially closed.
	$g^{(t)}$, is the input-update value. The model has two activation (squashing) units: input-update and output activation where the hyperbolic tangent $\mathrm{tanh}$ activation function is the preferable function to be used~\cite{elsayed2018a}.
	The memory cell state at time $t$ is $c^{(t)}$ and the output of the LSTM unit at time $t$ is $h^{(t)}$.
	
\begin{figure}
		\centering
		\includegraphics[width=8cm,height=5cm]{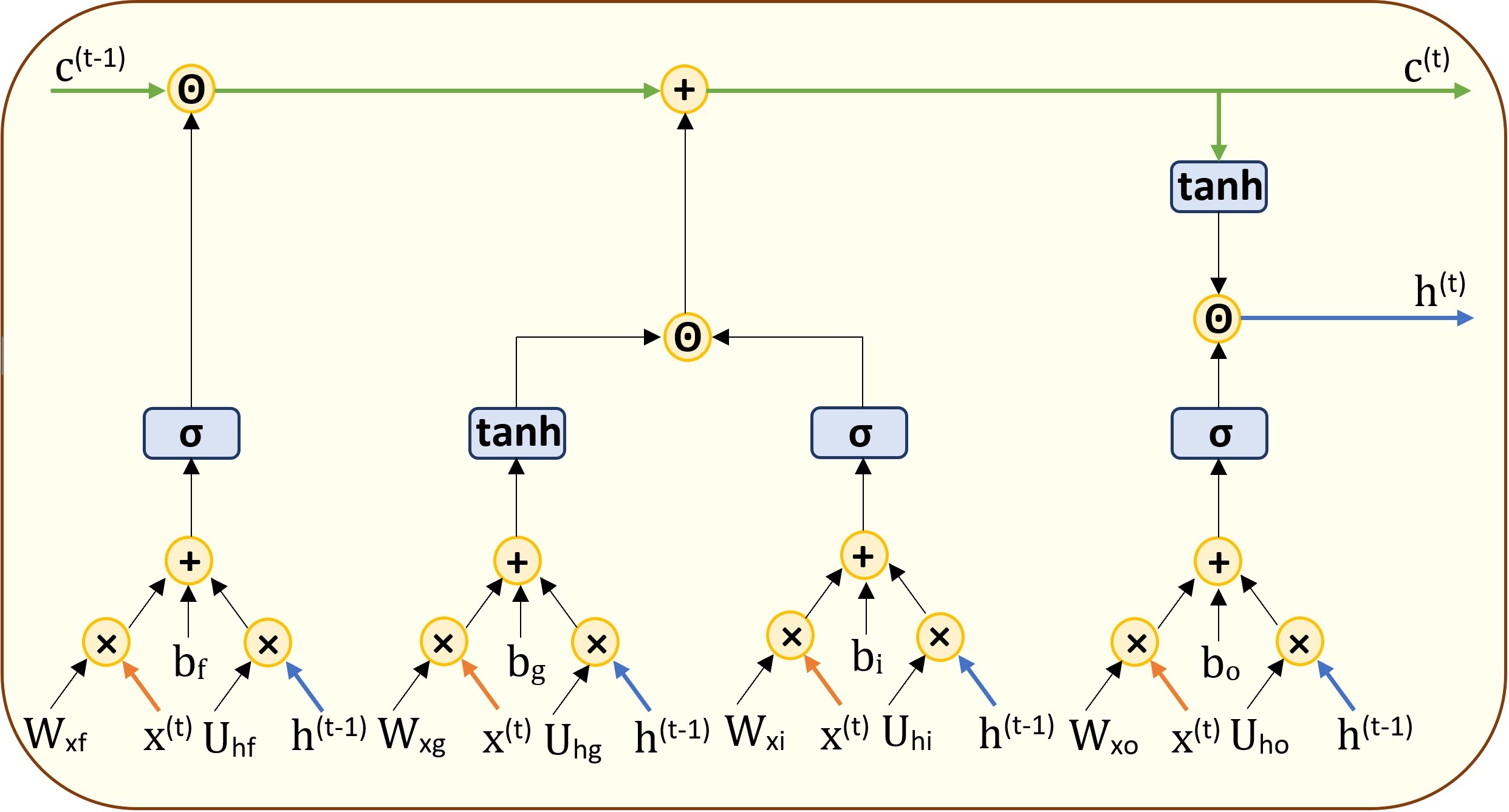}
		\caption{The standard LSTM unrolled architecture operation level that shows the components and their corresponding weights.} 
		\label{lstm_block_weights}
	\end{figure}

	Figure~\ref{lstm_block_weights} shows the operation level of the standard LSTM where each component of the standard LSTM and its corresponding weights are given. The symbols $\times$ and $\odot$ denote matrix multiplication and element-wise multiplication, respectively.
	
	The standard LSTM architectue is widely used in various problem-solving tasks and applications in different research fields. However, its architecture has major drawbacks. First, there is no direct connection from the memory to the gates which leads to the absence of CEC control over the gates~\cite{gers2002learning}. Second, if the output gate is closed, the CEC has no influence over the forget and input gates which could impair the model due to the lack of primary information flow within the model~\cite{gers2002learning}.
	\begin{figure}
		\centering
		\includegraphics[width=8cm,height=5cm]{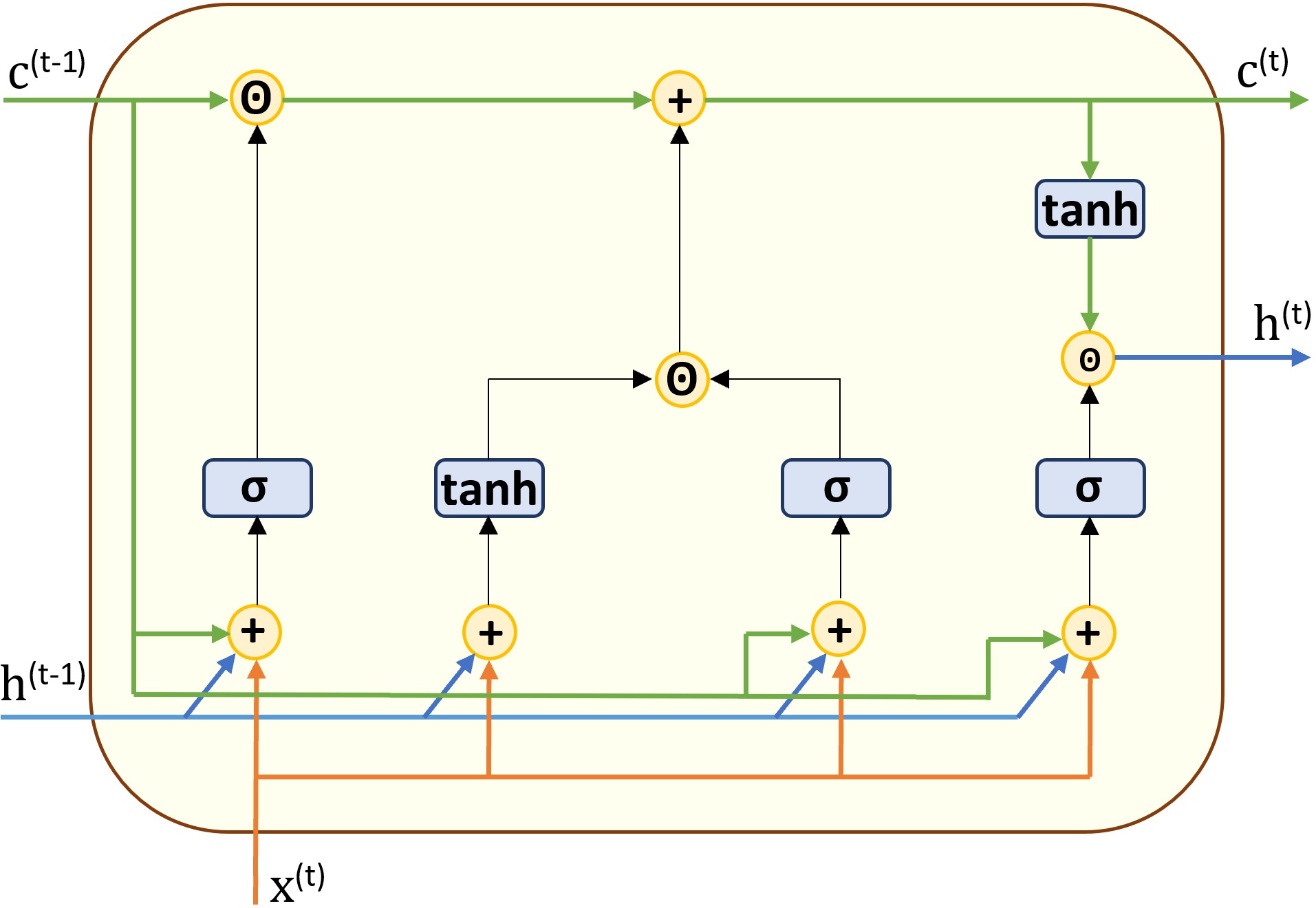}
		\caption{Gers et al.~\cite{gers2002learning} proposed peephole-based LSTM unrolled architecture.}
		\label{peephole_lstm_architecture}
	\end{figure}
	
	\subsection{The Peephole-Based LSTM}
	\label{peephole_lstm}

	
	Gers et al.~\cite{gers2002learning} proposed in 2002 a solution for the standard LSTM major problems. A new connection component has been added to the LSTM architecture named the peephole connection, in which data flow connection from the memory state to each of the three LSTM gates to solve the standard LSTM main problems. The peephole connections allow the memory state value to exert control over the LSTM three gates. This assists in preventing the vanishing and/or exploding gradient problem that the standard LSTM could face. 
	
\begin{figure}
		\centering
		\includegraphics[width=8cm,height=5cm]{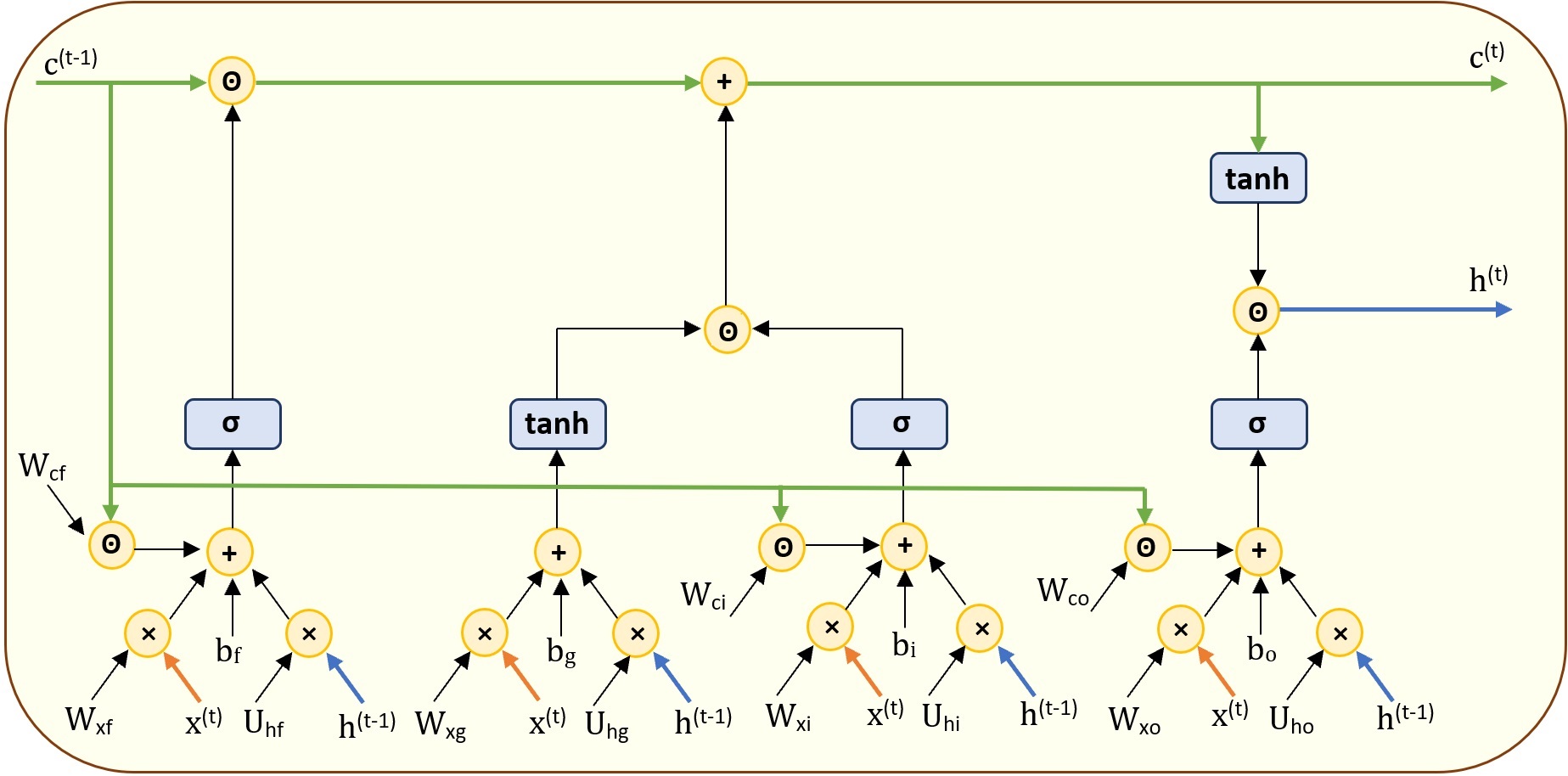}
		\caption{The operation level of the peephole-LSTM unrolled architecture where its components and their corresponding weights are presented.}
		\label{peephole_lstm_block_weights}
	\end{figure}

	Figure~\ref{peephole_lstm_block_weights} shows the operation level of the peephole-based LSTM. The equations to calculate the peephole LSTM are as follows:
	\begin{align}
		i^{(t)}&= \sigma(W_{xi} x^{(t)} +U_{hi}  h^{(t-1)}+ W_{si}\odot c^{(t-1)} + b_i)\label{eqn:i_plstm}\\ 
		g^{(t)}&= \mathrm{tanh}(W_{xg} x^{(t)} +U_{hg}  h^{(t-1)}+ b_g)\label{eqn:g_plstm}\\ 
		f^{(t)}&=\sigma(W_{xf} x^{(t)} +U_{hf}  h^{(t-1)} + W_{sf} \odot c^{(t-1)} + b_f)\label{eqn:f_plstm}\\ 
		o^{(t)}&=\sigma(W_{xo} x^{(t)} +U_{ho}  h^{(t-1)} + W_{so} \odot c^{(t-1)} + b_o)\label{eqn:o_plstm}\\ 
		c^{(t)}&= f^{(t)}\odot c^{(t-1)} + i^{(t)} \odot g^{(t)}\label{eqn:s_plstm}\\
		h^{(t)}&= \mathrm{tanh}(c^{(t)})\odot o^{(t)}\label{eqn:h_plstm}
	\end{align}
	where the symbol $\odot$ denotes the elementwise (Hadamard) multiplication. $W_{ci}$,  $W_{cf}$, and $W_{co}$ are the peephole connections weights between the memory state $c^{t-1}$ and the input, forget, and output gates, respectively.
	
	Adding the peephole connection to the standard LSTM made the LSTM architecture a robust model to overcome the vanishing and/or exploding gradient problem. However, it caused a significant increase in the number of trainable parameters, training time, and memory requirements. 
	
	\begin{figure}[h]
		\centering
		\includegraphics[width=8cm,height=5cm]{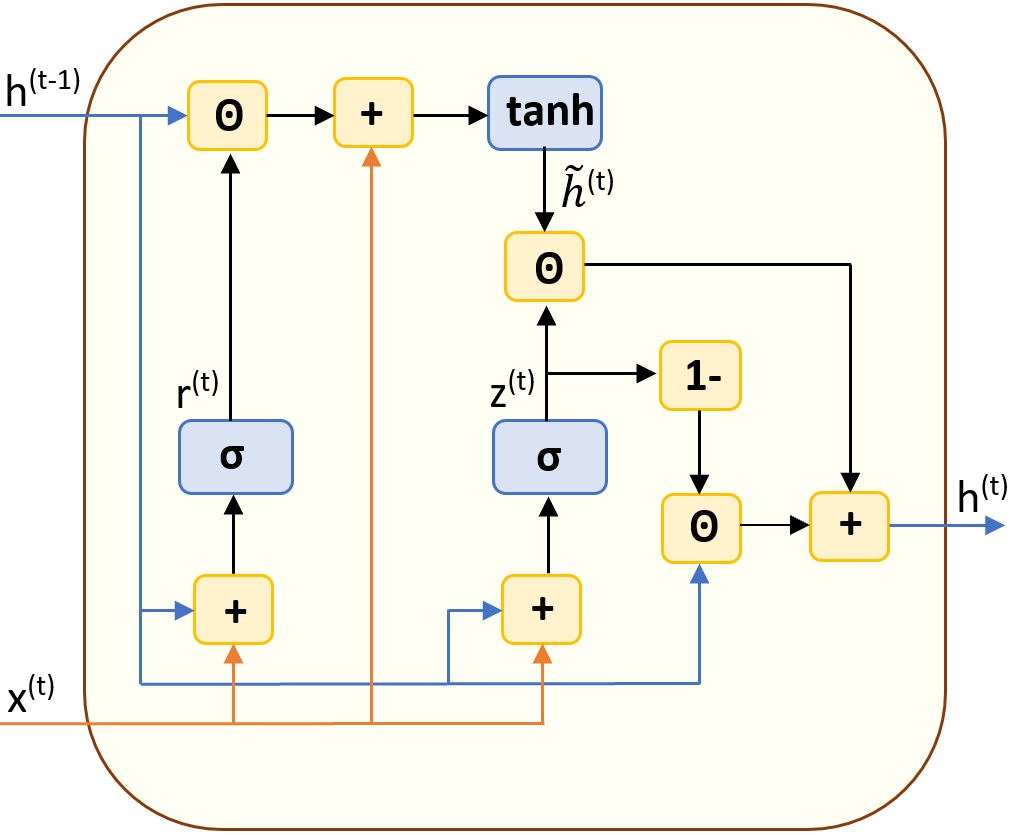}
		\caption{The GRU unfolder architecture.}
		\label{gru_block}
	\end{figure}
	
	\subsection{Gated Recurrent Unit (GRU)}\label{sec:GRU}

	The GRU model consists of two gates: the update gate $z$ and the reset gate $r$, whereas the LSTM consists of three gates: input, output, and forget gates. In addition, the GRU does not contain the memory state cell that the LSTM model includes. Therefore, the GRU architecture is smaller than the LSTM by one gate and a memory state cell. The GRU integrates both the input gate and forget gate of the LSTM model into one update gate $z$~\cite{greff2017lstm}, introducing the concept of the output of the same set of weights to reduce the model architecture. The unfolded GRU block architecture is shown in Figure~\ref{gru_block}.
	
	\begin{figure}
		\centering
		\includegraphics[width=8cm,height=5cm]{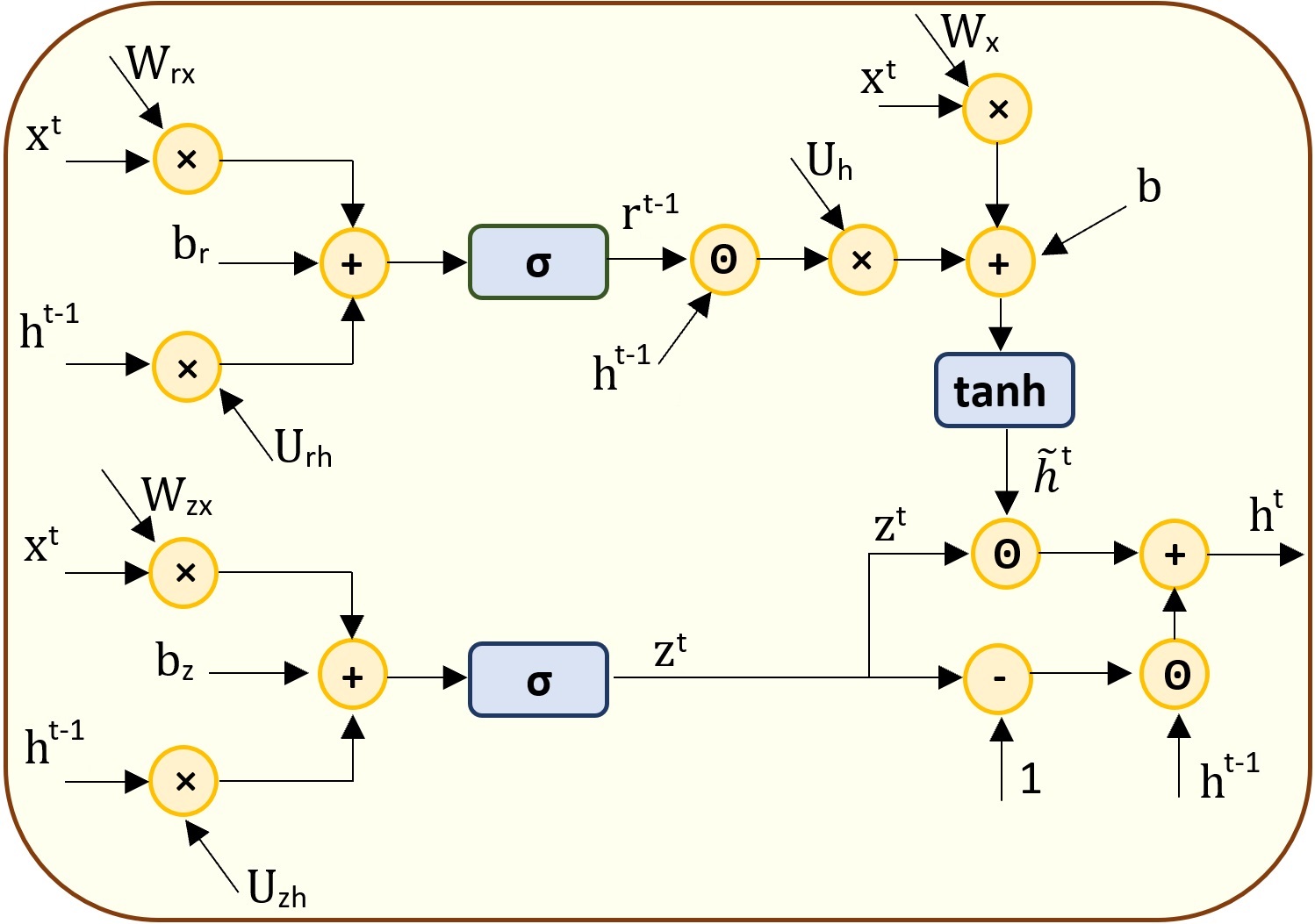}
		\caption{The operation level of the GRU architecture showing the weights of each component.}
		\label{gru_block_weights}
	\end{figure}	
	The reset gate functionality operates similarly to the output gate of the LSTM. 
	This GRU model eliminates the output squashing function, memory unit, and the CEC\@. 
	The GRU yields a reduction in trainable parameters compared with the standard LSTM.
	However, this may lead to exploding and/or vanishing gradients. 
	
	At time step, $t$\@, the GRU unit output, $h^{(t)}$, is calculated as follows~\cite{chung2014empirical}:
	
	\begin{align}
		z^{(t)} &= \sigma (W_{zx} x^{(t)} + U_{zh} h^{(t-1)}+ b_{z}) \label{eqn:z_gate_gru}\\
		r^{(t)} &= \sigma (W_{rx} x^{(t)} + U_{rh} h^{(t-1)}+b_{r})\label{eqn:r_gate_gru}\\
		{\tilde h}^{(t)} &= \mathrm{tanh} (W x^{(t)} + U( r^{(t)} \odot h^{(t-1)})+b)\label{eqn:h_hat_gru}\\
		h^{(t)} &= (1-z^{(t)})\odot h^{(t-1)}+z^{(t)}\odot{\tilde h}^{(t)}\label{eqn:h__gru}
	\end{align}
	\noindent
	where the $W_{xz}$, $W_{xr}$, and $W_{x}$ are the feedforward weights of the update gate $z^{(t)}$, the reset gate $r^{(t)}$, and the output candidate activation ${\tilde h}^{(t)}$, respectively. The recurrent weights are $U_{hz}$, $U_{hr}$, $U_{h}$ for the update gate $z^{(t)}$, the reset gate $r^{(t)}$, and the output candidate activation ${\tilde h}^{(t)}$, respectively. The biases of the update gate, reset gate, and the output candidate is denoted by $b_{z}$, $b_{r}$, and $b$, respectively. $\sigma$ is the logistic sigmoid function and $\mathrm{tanh}$ is the hyperbolic tangent function. The elementwise (Hadamard) multiplication is denoted by $\odot$. Figure~\ref{gru_block_weights} shows the operation level of the GRU architecture with weights and biases made explicit.
	
	\begin{figure}
		\centering
		\includegraphics[width=8cm,height=5cm]{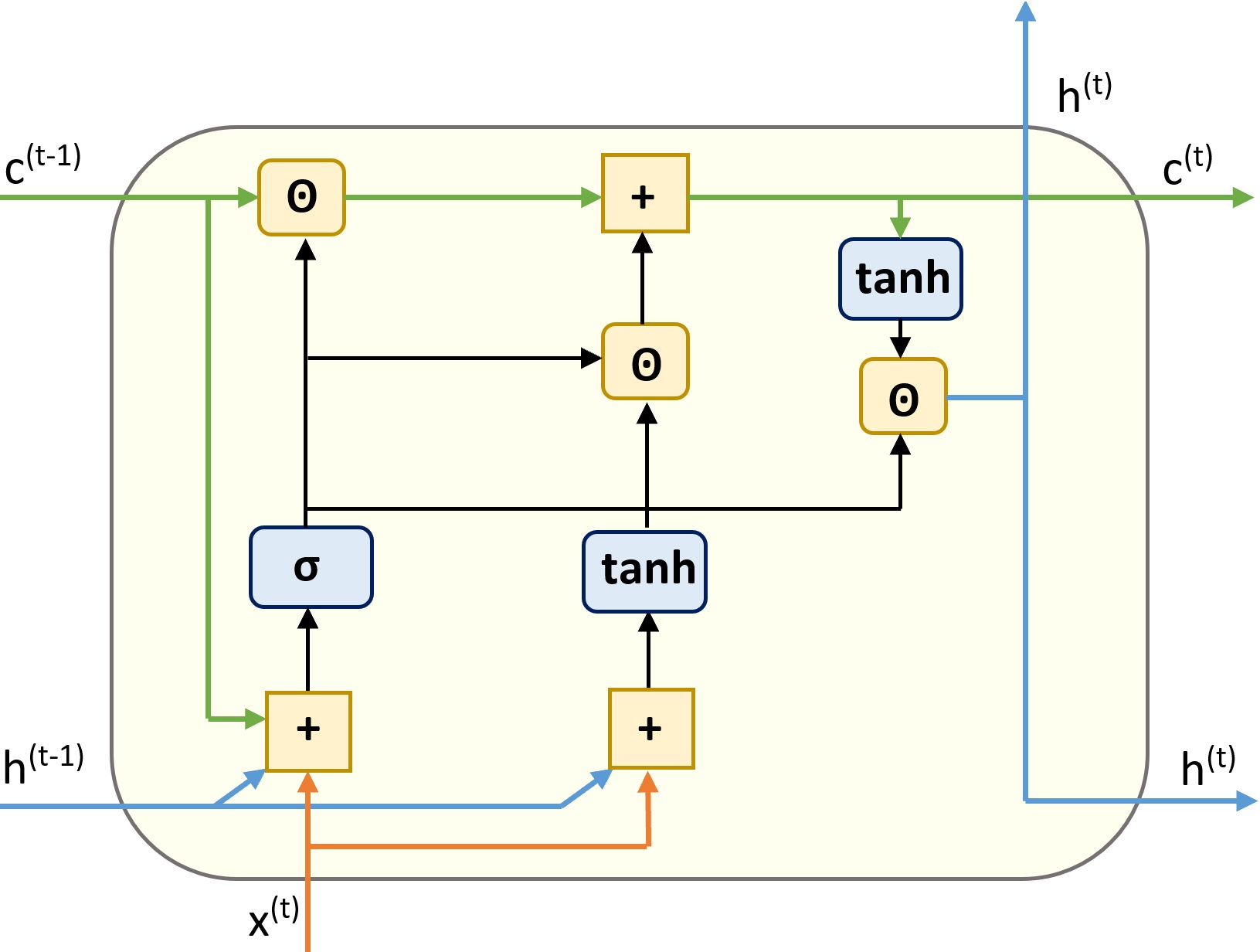}
		\caption{The LiteLSTM unrolled architecture. The single network gate (output indicated by $\sigma$)
			sends information flow to three locations that correspond to the outputs of the forget, input, and output gates of the standard LSTM.}
		\label{LiteLSTM_architecture}
	\end{figure}		
	\section{LiteLSTM Architecture}\label{LiteLSTM}
	The proposed LiteLSTM aims to: reduce the overall implementation cost of the LSTM, solve the LSTM significant problems, and maintain a comparable accuracy performance to the LSTM. The proposed LiteLSTM architecture appears in Figure~\ref{LiteLSTM_architecture}.
	
	The architecture of the LiteLSTM consists of only one trainable gated unit. We named the trainable gate the forget gate or network gate. This one gate behaves as a shared set of weights among the three gates of the standard LSTM gates. The LiteLSTM has a peephole connection from the memory state to the forget gate, which preserves the memory state from the LSTM and keeps the CEC to avoid vanishing and/or exploding gradients.
	
\begin{figure}
		\centering
		\includegraphics[width=8cm,height=5cm]{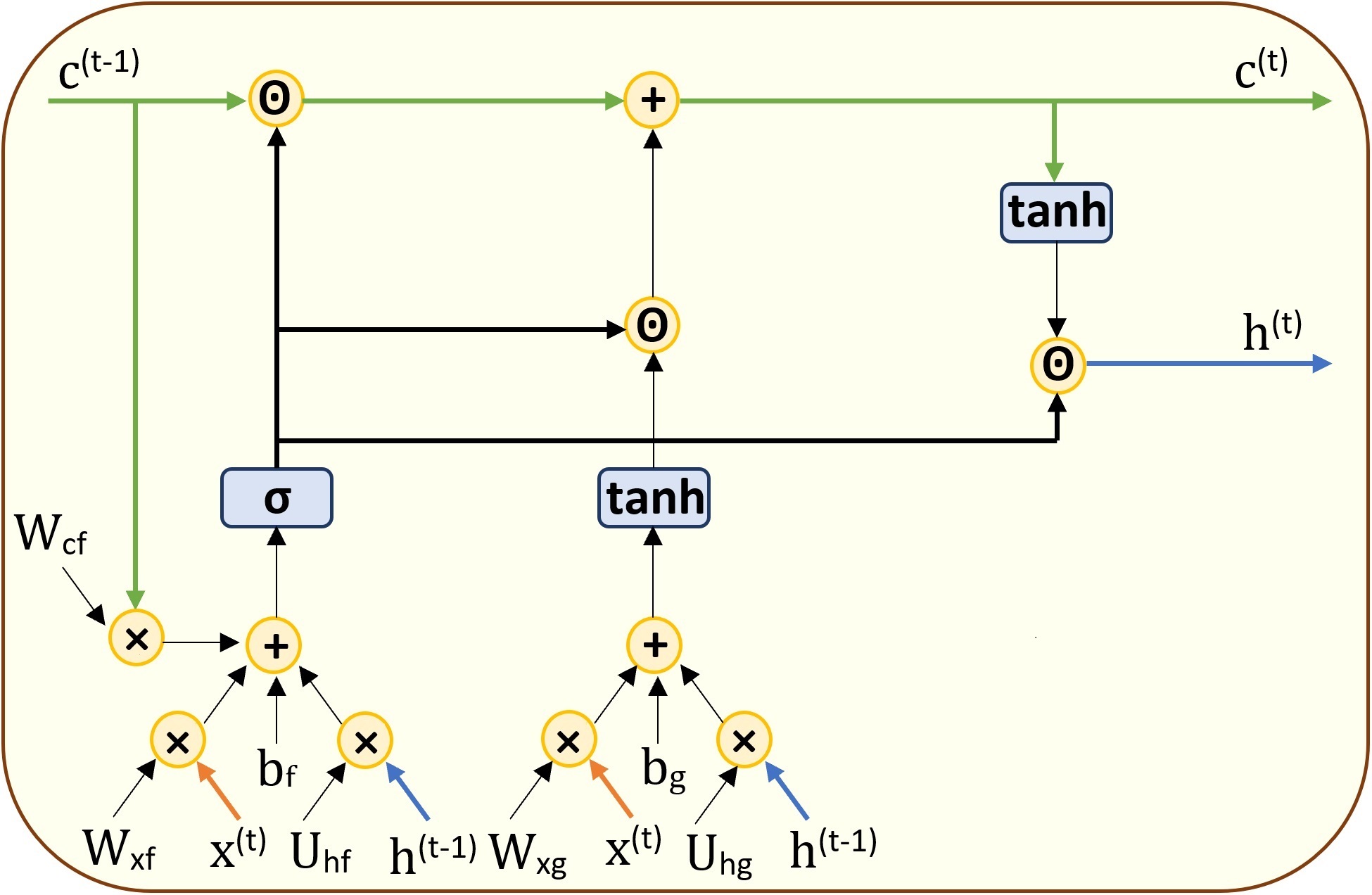}
		\caption{The operation level of the LiteLSTM architecture showing the weights of each component.}
		\label{LiteLSTM_weights}
	\end{figure}

	Thus, the proposed LiteLSTM preserves the critical components of the LSTM as stated by~\cite{greff2017lstm} while reducing much of the parameter redundancy in the LSTM architecture. 
	The LiteLSTM has a significant reduction in the number of trainable parameters that are required to implement the model. Therefore, the LiteLSTM reduced the training time, memory, and hardware requirements compared to the standard LSTM, peephole-based LSTM, and GRU architectures. 
	Furthermore, the proposed LiteLSTM architecture preserves comparable prediction accuracy results to the LSTM. Figure~\ref{LiteLSTM_weights} shows a detailed architecture of the unrolled (unfolded) LiteLSTM assuming non-stacked input.
	
	The LiteLSTM block architecture contains only one trainable gate that compensates the elimination of the other two gates of the standard LSTM by sharing its trainable weights. The LiteLSTM preserves the memory cell of the standard LSTM to process long data sequences and maintains the CEC to manage the vanishing/exploding gradient problem. 
	
	The LiteLSTM formulas are created as follows: During the forward pass within the LiteLSTM at time step $t$ the total input (inp), 
	$inp^{(t)}$, to the 
	single forget gate $f^{(t)}$ is calculated by:
	
	\begin{equation}\label{main_eqn}
		inp^{(t)} = \left[ W_{fx}, U_{fh}, W_{fc} \right] \left[x^{(t)}, h^{(t-1)}, c^{(t-1)}\right] + b_f
	\end{equation}
	
	\noindent
	where $inp^{(t)}\in\mathbb{R}^{\eta\times 1}$, and $\eta\times 1$ is the of input vector $inp^{(t)}$.
	$x^{(t)}$ is the input at time $t$, $ x^{(t)} \in \mathbb{R} ^{\eta  \times 1}$, $h^{(t-1)}$ is the output of the LiteLSTM architecture at time $t-1$, and the memory state cell at time $t-1$ denoted by $c^{(t-1)}$. Both $h^{(t-1)}, c^{(t-1)} \in \mathbb{R} ^{\eta  \times 1}$. $W_{fx}$, $U_{fh}$, and $W_{fc}$ are the weight sets.
	All three weight sets $W_{fx}$, $U_{fh}$, and $W_{fc}$ and biases $b_f$ are trainable. The square brackets indicate stacking. We will let $W_f = \left[ W_{fx}, U_{fh}, W_{fc} \right]$. 
	In addition, we let $I_f = \left[x^{(t)}, h^{(t-1)}, c^{(t-1)}\right]$.
	
	
	\begin{figure}
		\centering
		\includegraphics[width=7cm,height=4cm]{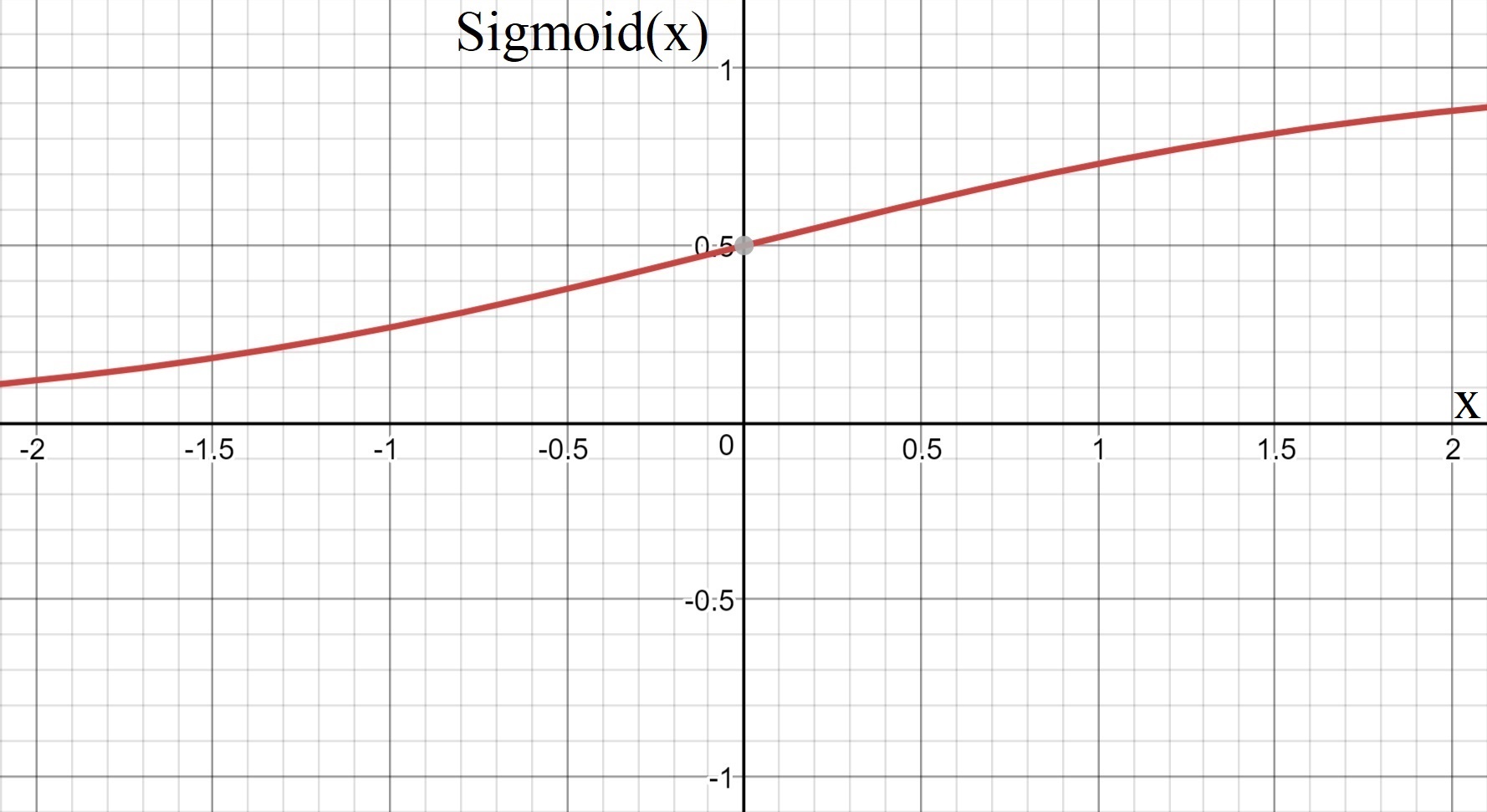}
		\caption{The logistic sigmoid function curve.}
		\label{sigmoid_fn}
	\end{figure}
	
	\begin{figure}
		\centering
		\includegraphics[width=7cm,height=4cm]{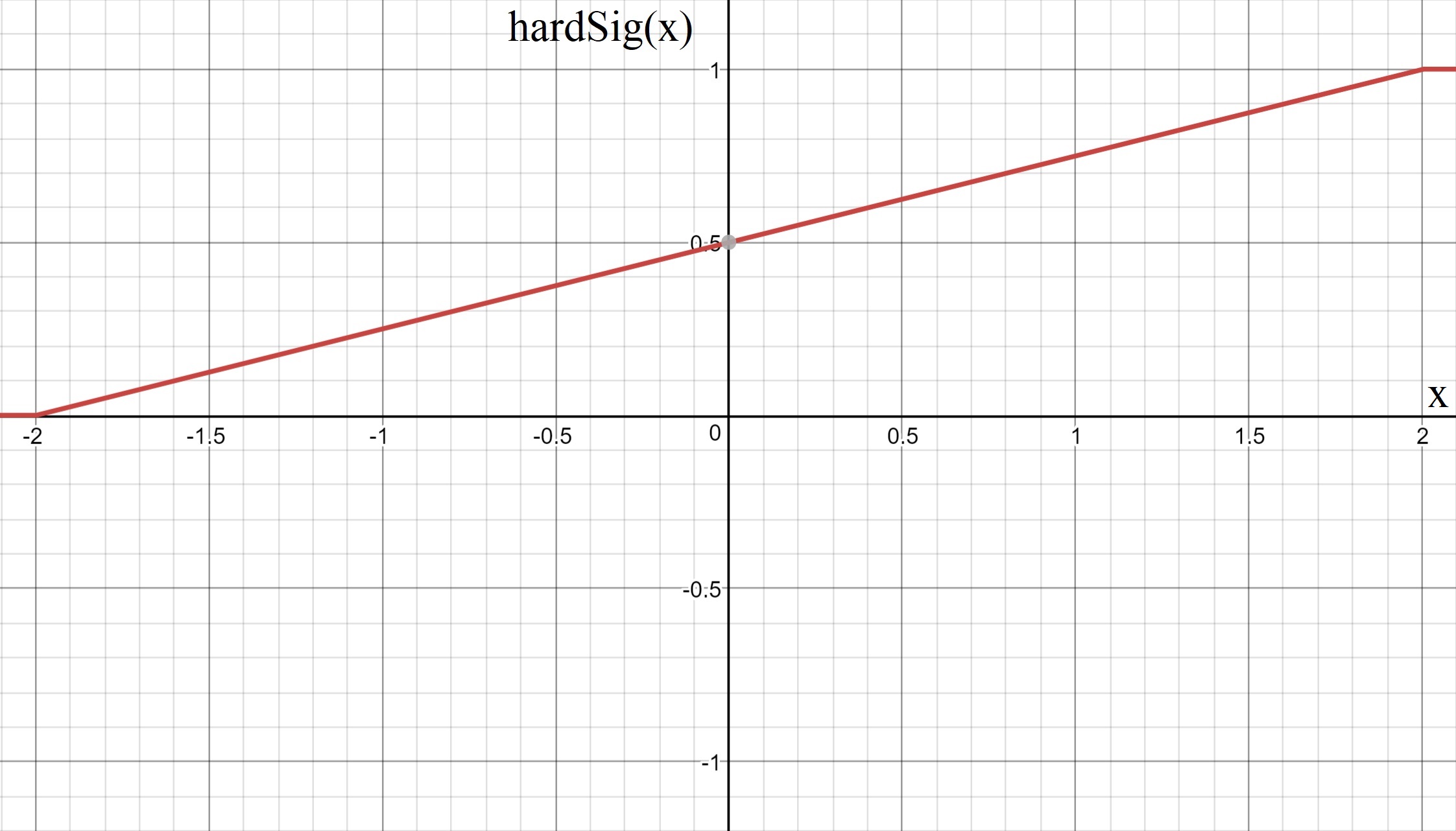}
		\caption{The hardSigmoid function curve.}
		\label{hard_sig}
	\end{figure}
	By applying a squashing function $G$ to the net input as follows:
	\begin{equation}\label{equx}
		f_\mathrm{gate}^{(t)}= G(inp^{(t)}) .
	\end{equation}
	Depending on the application, the squaching function $G$ can be either the logistic sigmoid ($\sigma$) or hard sigmoid ($\mathrm{hardSig}$)~\cite{Gulcehre2016}. The logistic sigmoid is calculated by:
	\begin{equation}\label{sigmoid}
		\sigma(x) = \frac{e^x}{e^x+1} = \frac{1}{1+e^{-x}} ,
	\end{equation}
	\noindent
	where $x$ is a real number, $x\in (-\infty,\infty)$, and $\sigma(x)$ has the range of $(0,1)$. The hard sigmoid ($\mathrm{hardSig}$) is calculated by: 
	\begin{equation}\label{hard_sigmoid}
		hardSig(x) = \max(\min(0.25x + 0.5, 1), 0) 
	\end{equation}
	\noindent
	Figure~\ref{sigmoid_fn} and Figure~\ref{hard_sig} shows the logistic sigmoid ($\sigma$) function and hard sigmoid ($\mathrm{hardSig}$) function curves, respectively. The values of $f^t$ in Eqn.~\ref{equx}
	falls in the range $(0,1)$ or $\left[0, 1\right]$, 
	depending on using the logistic sigmoid ($\sigma$) or hard sigmoid function, respectively~\cite{elsayed2018a,elsayed2019effects}. 	
	Assuming that case of selection the function as $\sigma$, the gate value $f^{t}$ is calculated by:
	\begin{equation}
		f^{(t)} = \sigma(W_{f}I_f + b_{f}) .
		\label{forget_gate_eqn}
	\end{equation}
	Selecting the logistic sigmoid or hard sigmoid functions is mainly based on the application. However, the hard sigmoid function is the preferred function to be used in the LiteLSTM gate to prevent the network gate from being closed (i.e., prevent the network gate from producing zero value output).
	The input update (memory activation) equation is calculated by:
	\begin{equation}
		\label{memoryActEqn}
		g^{(t)} =\tanh\left(W_g I_g + b_g\right)
	\end{equation}
	
	\noindent
where $W_g = \left[ W_{gx}, U_{gh} \right]$, 
	and $I_g = \left[x^{(t)}, h^{(t-1)}\right]$. The dimension in $W_g$ is matching the dimension of the $W_f$ that maintains the dimension compatability within the architecture design. 
	
	
	
	
	Finally, the Lite LSTM output is calculated by:
	
	\begin{align}
		c^{(t)} &= f^{(t)} \odot c^{(t-1)} + f^{(t)} \odot g^{(t)}\label{memory_cell}\\
		h^{(t)} &= f^{(t)} \odot tanh(c^{(t)})\label{rgcLSTM_output_eqn}
	\end{align}
	
	\noindent

	\begin{table*}
		\scriptsize 
		\setlength{\tabcolsep}{2.5pt}
		\caption{Computational components comparison between the proposed LiteLSTM and the state-of-the-art recurrent architectures.} 
		\centering 
		\begin{tabular}{ l  l l  l l l} 
			\textbf{Comparison} &\textbf{RNN}&\textbf{GRU} & \textbf{LSTM} &\textbf{pLSTM}&\textbf{LiteLSTM}\\
			\hline	\\
			Number of gates	& 0&	2	&	3& 3 & 1 \\
			Number of activations  &1&	1	&	2 & 2 & 2\\
			State memory cell & $\times$	&	$\times$	&	 $\checkmark$	&  $\checkmark$ & $\checkmark$\\
			Peephole connection& $\times$	& $\times$& $\times$& $\checkmark$& $\checkmark$ \\
			Number of weight matrices& 2& 6 & 8 &11 & 6\\
			Number of elementwise multiplication&2& 3 & 3 & 6 & 3\\
			Number of bias vectors&1& 3  &  4 & 4 & 2\\
	    	Sharing weights concept& $\times$	&	 $\checkmark$ &$\times$ & $\times$ & $\checkmark$\\
		\end{tabular}
		\label{comparison_tableable} 
	\end{table*}
	
	Table~\ref{comparison_tableable} shows a comparison between the architecture design and computation components of the RNN, GRU, standard LSTM, peephole-based LSTM (pLSTM), and the proposed LiteLSTM.
	
	\section{Emperical Evaluatuation and Analysis}\label{emperical}
	In this paper, the LiteLSTM has been empirically tested and evaluated in three research domains: computer vision, anomaly detection in IoT, and speech emotion recognition. The MNIST~\cite{lecun1998mnist} has been used as the computer vision experiment dataset, 
	and the IEEE IoT Network Intrusion Dataset~\cite{q70p-q449-19} is used for anomaly detection in IoT tasks. We used an Intel(R) Core(YM) i7-9700 CPU @3.00GHZ, 3000 Mhz processor, Microsoft Windows 10 OS, and 32 GB memory computer machine to perform our experiments. We used Python 3.7.6, Keras 2.0.4, and Tensorflow 1.15.0.
	
\begin{figure*}
		\centering
		\includegraphics[width=\textwidth,height=2.5cm]{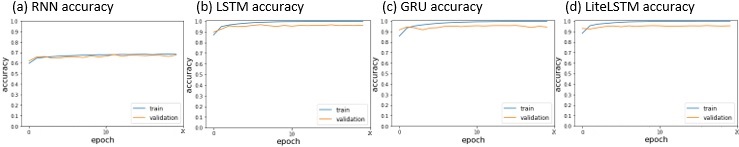}
		\caption{ The accuracy diagrams of the recurrent architectures and LiteLSTM using MNIST dataset.}
		\label{mnist_accuracy_plots}
	\end{figure*}

	The first empirical evaluation of the LiteLSTM was performed using the MNIST dataset, which consists of $70,000$ images of handwritten digits between 0 and 9. The dataset is split into $60,000$ data samples for training and $10,000$ data samples for testing~\cite{elsayed2022litelstm}. The MNIST images were centered in a 28$\times$28 image by computing the center of mass of the pixels. The model set 64-two layered architecture followed by a Softmax layer. For the training process, the batch size was set to 128 and the number of epochs to 20. The Adam optimizer with learning rate $10^{-3}$, $\beta_{1} = 0.9$, $\beta_{2}= 0.999$, and $\epsilon=1e-07$.  
	Table~\ref{mnist_result_Table} shows the accuracy results of the different recurrent architectures and the LiteLSTM, where the time is measured in minutes. The RNN shows a significantly shorter training time. However, it has the lowest performance compared to the other recurrent architectures. The LiteLSTM shows an improvement in accuracy compared to the other recurrent architectures. Figure~\ref{mnist_accuracy_plots} shows the accuracy plots for each of the LiteLSTM and the state-of-the-art recurrent models. 
	
	\begin{table}
		\setlength{\tabcolsep}{2.5pt}
		\caption{Accuracy comparision between the LiteLSTM and the state-of-the-art recurrent architectures using MNIST dataset} 
		\centering 
		\begin{tabular}{ l  l  ll ll} 
			\textbf{Comparision} & \textbf{RNN}& \textbf{GRU}&\textbf{LSTM}&\textbf{pLSTM}&\textbf{LiteLSTM} \\
			\hline	\\
			\textbf{Time(m)}& \textbf{11.24}& \textbf{43.01}& 60.36& 75.45& \textbf{42.94}\\
			\textbf{Parameters}& \textbf{792,210}&812,610&822,810& 833,010& 812,610 \\ 
			\textbf{Accuracy(\%)}& 67.64\%& 94.09\% &95.70\%&95.99\%&\textbf{96.07\%}\\
			
		\end{tabular}
		\label{mnist_result_Table} 
	\end{table}
	
	The second empirical evaluation of the LiteLSTM was performed using the IEEE IoT Network Intrusion Dataset. The dataset consists of 42 raw network packet files (pcap) at different time points. The IoT devices, namely SKT NUGU (NU 100) and EZVIZ Wi-Fi camera (C2C Mini O Plus 1080P) were used to generate traffic for IoT devices. The data contains normal traffic flow and different types of cyberattacks, namely: ARP spoofing attack, DoS (SYN flooding) attack, scan (host and port scan) attack, scan(port and OS scan) attack, (UDP/ACK/HTTP Flooding) of zombie PC compromised by Mirai malware, Mirai-ACK flooding attack, Mirai-HTTP flooding attack, and Telnet brute-force attack. In our experiments, we used a dataset to experiment with the LiteLSTM twice: first, to detect whether an attack occurred or not (as a binary dataset), and another experiment to detect the type of attack. We set the batch size to 32 and the number of epochs to 20. Table~\ref{IoT_results_binary_table} shows the binary experimental results for the LiteLSTM and the recurrent architectures. Table~\ref{IoTmulti_results} shows the detection results of the LiteLSTM and the recurrent architectures for detecting different types of cyberattacks.
	
	\begin{table}
		\setlength{\tabcolsep}{2.5pt}
		\caption{Accuracy comparision between the LiteLSTM and the state-of-the-art recurrent architectures using IEEE IoT Network Intrusion Binary Dataset} 
		\centering 
		\begin{tabular}{ l  l  ll  l l} 
			\textbf{Comparison} &\textbf{RNN}&\textbf{GRU}&\textbf{LSTM}&\textbf{pLSTM}&\textbf{LiteLSTM} \\
			\hline	\\
			\textbf{Time (m)} & \textbf{20.26}&43.27 & 41.51& 51.21& 28.44 \\
			\textbf{Precision}& 0.8144&0.9328 & 0.9422 & \textbf{0.9653} & 0.9382 \\ 
			\textbf{Recall}& 0.9763&0.9757& 0.9484 & 0.9545 &\textbf{0.9834}\\
			\textbf{F1-score} & 88.80&91.34 & 95.97& 95.99&\textbf{0.9603}\\
			\textbf{Accuracy(\%)}& 98.7\%&99.51\% & 99.50\% & 99.56\% & \textbf{99.60\%}\\
			
		\end{tabular}
		\label{IoT_results_binary_table} 
	\end{table}

	\begin{table}
		\setlength{\tabcolsep}{2.5pt}
		\caption{Accuracy comparison between the LiteLSTM and the state-of-the-art recurrent architectures using IEEE IoT Network Intrusion Detection for Multiple Classes Cyberattacks Dataset.} 
		\centering 
		\begin{tabular}{ l  l  ll  l l} 
			\textbf{Comparison} &\textbf{RNN}&\textbf{GRU}&\textbf{LSTM}&\textbf{pLSTM}&\textbf{LiteLSTM} \\
			\hline	\\
			\textbf{Time (m) }&  \textbf{19.98}  & 42.79     & 50.41     &  59.96      &  29.31 \\
			\textbf{Precision}&          0.8875   & 0.8991    & 0.9461   &0.9249      & 0.8999 \\ 
			\textbf{Recall}&             0.8418   & 0.8300    & 0.7898    & 0.8086       & \textbf{0.8318}\\
			\textbf{F1-score}&           0.8640   & 0.8632    & 0.8609    & 0.8628      & \textbf{0.8645}\\
			\textbf{Accuracy(\%)}&       83.35\% & 86.70\%  & 86.90\%  & 87.03\%     & \textbf{87.10\%} \\
			
		\end{tabular}
		\label{IoTmulti_results} 
	\end{table}
	
	\begin{figure*}
		\centering
		\includegraphics[width=\textwidth,height=2.5cm]{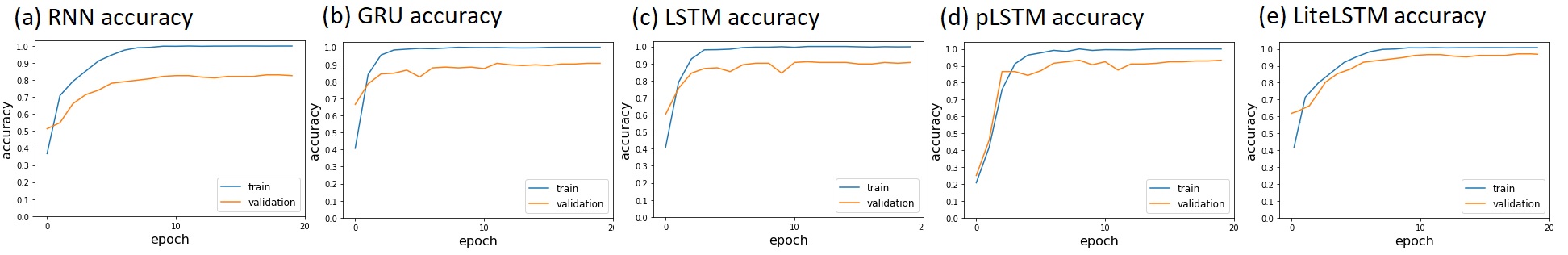}
		\caption{The accuracy diagrams of the recurrent architectures and LiteLSTM using Toronto Emotion Speech Set (TESS) dataset.}
		\label{tess_accuracy_plot}
	\end{figure*}

	The third empirical evaluation of the LiteLSTM was performed on a voice (audio) emotion recognition task. For this purpose, we used the Toronto Emotional Speech Set (TESS)~\cite{dupuis2010toronto}, which is one of the emotion recognition dataset benchmarks that has been used in several emotion recognition applications and tasks~\cite{elsayed2022speech,gokilavani2022ravdness,parry2019analysis}. This dataset consists of 2800 stimuli and has seven different emotion categories: anger, disgust, fear, happiness, pleasant/surprise, sadness, and neutral. The major significance of this dataset is that the distribution between the number of stimuli per emotion category is equally likely~\cite{dupuis2010toronto}. Similar to the previous experiments, we tested the proposed LiteLSTM with the other recurrent neural network architectures. For this empirical evaluation, we used the model described~\cite{elsayed2022speech}, which used the GRU as the learning model. We replaced the GRU with LiteLSTM, peephole LSTM, and RNN and evaluated the model performance each time. The dataset has been split into training, testing, and validation sets with a ratio of 70\%, 20\%, and 10\%, respectively. Table~\ref{tess_results} shows the empirical result of the proposed LiteLSTM and the recurrent architectures for emotion recognition from speech. Figure~\ref{tess_accuracy_plot} shows the training versus validation accuracies for each of the recurrent architectures and LiteLSTM using Toronto Emotion Speech Set (TESS) dataset.

	\begin{table}

		\setlength{\tabcolsep}{2.5pt}
		\caption{Accuracy comparison between the LiteLSTM and the state-of-the-art recurrent architectures using the Toronto Emotional Speech Set (TESS).} 
		\centering 
		\begin{tabular}{ l  l  ll  l l} 
			\textbf{Comparison} &\textbf{RNN}&\textbf{GRU}&\textbf{LSTM}&\textbf{pLSTM}&\textbf{LiteLSTM} \\
			\hline	\\
			\textbf{Time (m) }& 	79.56&171.16&201.64&239.84&117.24\\
			\textbf{Precision}& 0.9312   &   0.9428 &   0.9686 &     \textbf{0.9898}  & 0.9799 \\ 
			\textbf{Recall}&    0.9546    &   0.9429 &  0.9026   &     0.9214   & \textbf{0.9446}\\
			\textbf{F1-score}& 0.9427      &   0.9428 &   0.9344    &   0.9543 &\textbf{0.9619}\\
			\textbf{Accuracy(\%)}&       92.163\%  &   94.285\% & 95.147\% &   95.534\%  & \textbf{95.989\%}  \\
			
		\end{tabular}
		\label{tess_results} 

	\end{table}

	\section{Conclusion}
	The proposed LiteLSTM architecture novelty lies in the following aspects. First, the LiteLSTM consists of one gate that serves as a multifunctional gate via the weights-sharing concept. Thus, the overall number of training parameters is reduced by approximately one-third of the LSTM or the peephole-LSTM. In addition, maintaining the peephole connection from the memory state cell to the existing gate maintains the control of the memory over the gate in contrast to the LSTM. Therefore, the LiteLSTM handles the vanishing/exploding gradient problem.The overall budget for implementing the LiteLSTM, including the training time, memory footprint, memory storage, and processing power, is smaller than the LSTM by approximately one-third. We empirically evaluated the LiteLSTM using three datasets: MNIST, IEEE IoT Network Intrusion Detection datasets, and TESS speech emotion recognition dataset. The proposed LiteLSTM shows comparable results to the LSTM using a smaller computation budget. Due to the optimized LiteLSTM architecture design, we were able to complete the empirical tasks using a computer processor without involving the GPU in the computational process. Thus, the LiteLSTM architecture helps to reduce the CO2 footprint. The proposed LiteLSTM architecture is an attractive candidate for future hardware implementation on small and portable devices, especially IoT devices.
	
	\section*{Statements and Declarations}
	
	
	\begin{itemize}
		\item \textbf{Funding:} N/A
		\item \textbf{Conflict of interest/Competing interests:} The authors declare that they have no conflict of interest.	
		\item The authors did not receive support from any organization for the submitted work.
		\item All authors certify that they have no affiliations with or involvement in any organization or entity with any financial interest or non-financial interest in the subject matter or materials discussed in this manuscript.
		\item The authors have no financial or proprietary interests in any material discussed in this article.
	\end{itemize}
	\bibliographystyle{sn-mathphys}
	\bibliography{sn-bibliography}
	
	
\end{document}